\documentclass[conference]{IEEEtran}
\IEEEoverridecommandlockouts
\usepackage{cite}
\usepackage{amsmath,amssymb,amsfonts}
\usepackage{algorithmic}
\usepackage{graphicx}
\usepackage{textcomp}
\usepackage{subfigure} 
\usepackage{multirow}
\usepackage{xcolor}
\def\BibTeX{{\rm B\kern-.05em{\sc i\kern-.025em b}\kern-.08em
    T\kern-.1667em\lower.7ex\hbox{E}\kern-.125emX}}
\begin{document}

\title{A Chit-Chats Enhanced Task-Oriented Dialogue \\ Corpora for Fuse-Motive Conversation Systems}

\author{\IEEEauthorblockN{1\textsuperscript{st} Changhong Yu}
\IEEEauthorblockA{\textit{Beijing University of Posts}\\
\textit{and Telecommunications}\\
Beijing, China \\
charpyu@bupt.edu.cn}
\and
\IEEEauthorblockN{2\textsuperscript{nd} Chunhong Zhang}
\IEEEauthorblockA{\textit{Beijing University of Posts}\\
\textit{and Telecommunications}\\
Beijing, China \\
zhangch@bupt.edu.cn}
\and
\IEEEauthorblockN{3\textsuperscript{rd} Qi Sun}
\IEEEauthorblockA{\textit{Beijing University of Posts}\\
\textit{and Telecommunications}\\
Beijing, China \\
qisun@bupt.edu.cn}
}

\maketitle

\begin{abstract}
The goal of building intelligent dialogue systems has largely been separately pursued under two motives: task-oriented dialogue (TOD) systems, and open-domain systems for chit-chats (CC). Although previous TOD dialogue systems work well in the testing sets of benchmarks, they would lead to undesirable failure when being exposed to natural scenarios in practice, where user utterances can be of high motive-diversity that fusing both TOD and CC in multi-turn interaction. 
Since an industrial TOD system should be able to converse with the user between TOD and CC motives, constructing a fuse-motive dialogue dataset that contains both TOD or CC is important. Most prior work relies on crowd workers to collect and annotate large scale dataset and is restricted to English language settings. Our work, on the contrary, addresses this problem in a more effective way and releases a multi-turn dialogues
dataset called Chinese Chat-Enhanced-Task (CCET). Meanwhile, we also propose a line of fuse-motive dialogues formalization approach, along with several evaluation metrics for TOD sessions that are integrated by CC utterances.
\end{abstract}

\begin{IEEEkeywords}
dialogue dataset, fuse-motive, task-oriented dialogue (TOD), chit-chats (CC)
\end{IEEEkeywords}

\section{Introduction}
Building intelligent dialogue systems (e.g., \cite{wen2016network,lei2018sequicity,bocklisch2017rasa,lee2019convlab}) has advanced significantly in recent years.
In terms of applications, dialogue systems are commonly classiﬁed into two categories: task-oriented dialogue (TOD) systems\cite{wang2020large,wang2021naturalconv,qin2021gl} and open-domain systems for chit-chats (CC)\cite{zhao2017generative}. TOD systems perform goal-oriented functions to achieve certain task objectives, while CC systems make chit-chats as a human speaker.

Generally, most previous works on TOD systems focus on making research efforts to improve the performance \cite{kim2021oh} on public TOD benchmarks. Actually, in real-life conversation\cite{ni2021recent}, the user utterances usually have the diversity of grammatical expression, language variations and dialogue scenario. This diversity leads to the unfortune that the TOD systems would be vulnerable when users switch their input motive from certain in-domain goals to open-domain chit-chats, which are not observed in training data. And most public TOD datasets \cite{ye2021multiwoz} that collected and annotated by disciplined crowdworkers can not simulate real-life conversation interaction. 
In short, it's vital to put TOD systems in place that can respond to non-distributed user input in real-world dialogues.

Previous researches \cite{madotto2020adapter, yan2017building} on this issue can be categorized into two types: model-fused and data-fused. 
The model-fused approaches, intuitively, learn TOD and CC conversations with separate models on different datasets. A selector or gating function is usually built above separate models to determine the response to the user per turn. Compared to a single model, the fused-model performs on a large scale of parameters, making algorithm optimization computationally expensive. The data-fused approaches, on the contrary, aim to create a multi-turn benchmark dataset that contains both TOD and CC types of utterances with inter-turn contextual dependency. In this way, a direct solution would be to have a single unified end-to-end architecture for fuse-motive dialogue. A few work have studied different methods to collect and construct fuse-motive dialogue datasets relying on crowd workers. However, it is usually challenging and costly to collect and build high-quality corpus in multi-turn conversation settings. And for each conversational turn in TOD, it is tedious to label several relevant annotations such as user intents, belief states by crowed workers. Therefore, it is intuitive to create a fused dialogue datasets automatically. Actually, this line of thought is challenged by two issues. First, the corpus generated with the help of rule-based dialogue simulators is still insoluble to displace real-world user utterances. Second, it is hard to construct a high-quality fused dataset automatically since annotation strategys for data of TOD and CC types are markedly different.

In this paper, we follow the dataset-fused thread but efficiently construct a fuse-motive dialogue dataset in an automatic way.
Based on several heterogeneous public Chinese datasets, including RisaWOZ\cite{quan2020risawoz}, LCCC-base\cite{wang2020large}, and NaturalConv\cite{wang2021naturalconv}, we work with alignment and fusion operations to create a multi-turn conversation dataset called \textbf{C}hinese \textbf{C}hat \textbf{E}nhanced \textbf{T}OD (CCET). We analyze the strengths and weaknesses of our dataset:
\begin{itemize}
	\item CCET contains both TOD and CC types of utterances, which can mimic real-life interaction environments.
	\item CCET could enable existing architecture of TOD models to possess generic chatting abilities of CC dialogue systems with no annotation cost and thus readily access in industry. 
	\item CCET might has a lower quality compared with the elaborately-collected human data because it is constructed automatically.
\end{itemize}

Our main contributions are as follows: 
\begin{itemize}
	\item We propose an effective construction approach to add CC interactions to TOD benchmarks automatically.
	\item We release a multi-turn fuse-motive dialogue dataset, CCET, as an instance of the above approach in the Chinese language setting.
	\item Along with CCET, a model-agnostic paradigm of building TOD systems that integrate CC interaction skills is also provided. Any baselines of TOD could be enhanced with the capability to handle casual talk by using our CCET dataset.
\end{itemize}
The CCET dataset and paradigm code are freely available under https://github.com/HunYuanFeng/CCET2021.

\section{Related Work}
The development of task-oriented dialogue (TOD) systems has been consistently supported by the appearance of human-created datasets\cite{adiwardana2020towards,budzianowski2018multiwoz,rastogi2020towards}. However, the distribution of user intents in these datasets are typically unmatched by the diversity in practice, where casual chit-chats (CC) could usually leads disfluency. The models trained on these corpora would be vulnerable or misled when exposed to realistic data distribution.

Previous works attempt to tackle this issue in two different ways: model-fused, or data-fused. The former approach is implemented as a modularized framework \cite{madotto2020adapter}, consists a task-oriented system, a chit-chats chatbot, and a mode-selector. In this way, different types of dialogue are assigned to independent models. The latter line of works, on the contrary, converses seamlessly and naturally in not only TODs but also casual chit-chats talk through a united architecture. 

It appeals to a model-agnostic multi-turn dialogue corpora that contains rich annotations \cite{madotto2020adapter} for both TOD and CC goals. \cite{sun2020adding} first use GPT2-based chatbots to generate CC candidates by conditioning on TOD context, then ask human annotators to edit candidates with justifications. These created CC supplements are added to TOD corpora to build fuse-motive dataset. 
\cite{wang2021naturalconv} collect a CC dataset that involves some specific topics such as movies, which could be utilized as the context of in-domain TOD interaction. However, Most prior work \cite{ni2021recent} relies on crowd workers to create large scale multi-turn dataset, which requires high annotation cost and thus are hard and expensive to access in industry.
Our released CCET, innovatively, targets the issue of integrating chitchats into TODs utterances in a effective and affordable manner with the help of existing publicized datasets. Based on three Chinese open-source datasets, RisaWOZ\cite{quan2020risawoz}, LCCC-base\cite{wang2020large}, and NaturalConv\cite{wang2021naturalconv}, we propose a methodology to transform traditional TOD annotation structure to a novel formulation, so that the CC type utterances could be accommodated. Then, we automatically align and fuse the TOD and CC types of samples together to mimic user inputs in real-world.

\section{Overview of Public Datasets} 
This section introduces the three utilized datasets in our work respectively. Table \ref{table:NaturalConv} exhibits some dialogue examples excerpted from conversations in these datasets.

\subsection{RisaWOZ}
RisaWOZ \cite{quan2020risawoz} is a task-oriented multi-domain Chinese dataset, ranging 12 domains, namely Attraction, Restaurant, Hotel, Flight, Train, Weather, Movie, TV, Computer, Car, Hospital and Education, etc. Among them, sessions in several domains, such as Hotel, Flight and Weather, triggers mix dialogue skills in terms of more complex tasks such as making travel plans. While sessions in some other domains, like TV, Computer and Car, focus on purchase-related intents with paraphrases of product-related
terms.

\begin{table*}[htbp]
\caption{Examples of three conversation within RisaWOZ, LCCC-base and NaturalConv datasets.}
\begin{center}
\boldmath
\scalebox{0.75}{%
	\begin{tabular}{p{7cm}|p{7cm}|p{7cm}}
	\hline
{\textbf{RisaWoz}} & {\textbf{LCCC-base}} & {\textbf{NeuralConv}} \\ \hline
\texttt{User}: Hello, I am a tourist in Suzhou. Could you recommend a funny place for me? I'm here with a friend so please find a place for friends. 

\texttt{System}: Which area would you like to look for attractions in. 

\texttt{User}: I'd like to find something in Kunshan at a medium price. 

\texttt{System}: The Jinxi Ancient Town is recommended for you. 

\texttt{User}: Ok, let's go to this spot. When is it open? My friends and I would book a time in advance. 

\texttt{System}: It's open from 8 a.m. to 5 p.m. 

&

\texttt{Speaker1}: You don’t need sleep? 

\texttt{Speaker2}: I lose sleep last night. 

\texttt{Speaker1}: What have you done?

\texttt{Speaker2}: Probably I am too excited for holidays. 

\texttt{Speaker1}: Okay!

&
\texttt{Speaker1}:  I've been looking forward to this movie for a long time since the trailer.

\texttt{Speaker2}:  Actually I don't really like cop movies, but this one is really good.

\texttt{Speaker1}:  I heard that Yao Yao has won an award for this movie.

\texttt{Speaker2}:  As a newcomer, she acts really good already.

\texttt{Speaker1}:  Have you seen \textit{Drug Dealer}?

\texttt{Speaker2}:  Yes, I did. I couldn't miss such a good movie.\\

\hline
\end{tabular}
}

\label{table:NaturalConv}
\end{center}
\end{table*}

The dataset is the largest publicized Chinese TOD dialogue dataset so far, containing about 11.2K multi-turn dialogue sessions, totally over 150K turns of user \& system interaction utterances. Apart from collecting the utterances, a group of domain experts are asked to develop a structured ontology for TOD systems to define the slot and possible values for specific domains. In RisaWOZ, different slots are identified as the \textit{Informable Slots}, referring to the slots where users can provide information, and the \textit{Requestable Slots}, where users can ask for values to back-end databases, under each domain within RisaWOZ. Then, crowd workers annotate the user goals, belief states, system actions and responses for the whole dataset. In this paper, these annotations would be transformed to accommodate chit-chats corpora.

\subsection{LCCC-base}
LCCC \cite{wang2020large} is a large-scale cleaned Chinese conversation dataset presented for developing open-domain chatbots. It contains a base version (6.8 million dialogues) and a large version (12.0 million dialogues) and of a high quality. In our work, CCET mainly utilizes utterances in this corpora to enhance the interaction for general TOD systems.

\subsection{NaturalConv} 
NaturalConv\cite{wang2021naturalconv} is a multi-turn CC dataset provided by Tencent AI Lab, containing about 400K utterances and 19.9K sessions across multiple topics like Sport, Entertainment and Technology. Different from LCCC-base, which are mainly open-domain short-text utterances, NaturalConv has more specific topics to embrace long-text sessions with more complex logic. The average number of turns is 20 \cite{wang2021naturalconv}, which is significantly longer than other released Chinese corpora. NaturalConv features topic-driven Chinese conversation generation, concentrates on conversation with social attributes, such as scenario adaption, free topic expansion, greetings, etc. CCET adds NaturalConv to LCCC-base to increase the relevancy to domain-related terms of chit-chats utterances like TOD interaction. 

\begin{figure*}[htbp]
\centerline{\includegraphics[width=1.6\columnwidth]{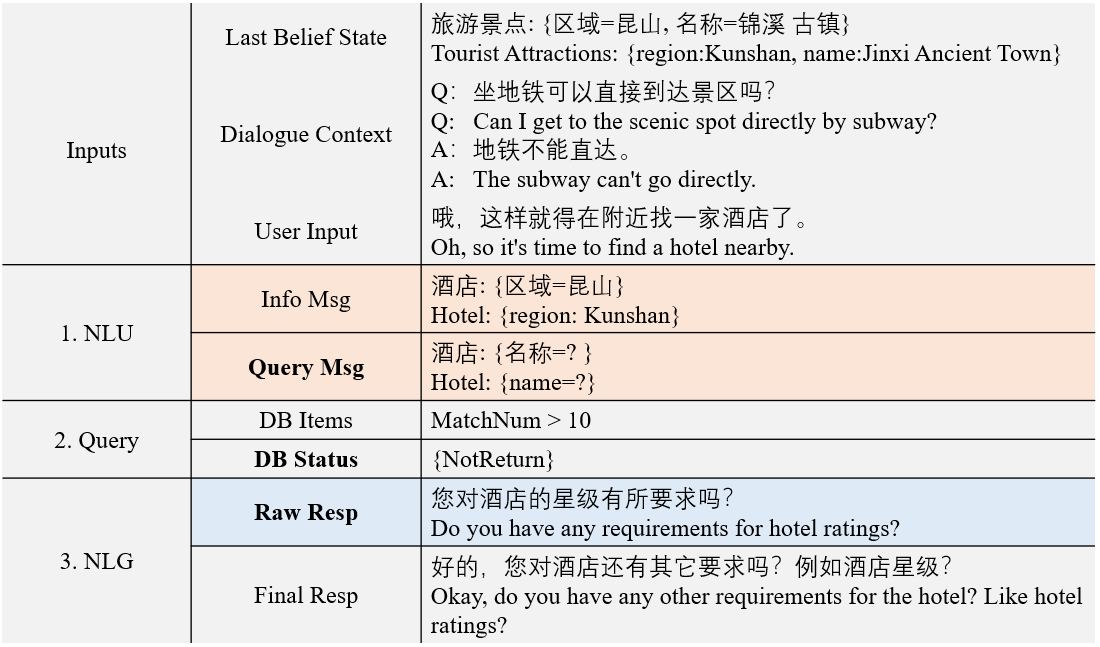}}
\caption{The data structure of a turn in the chat-enhanced-task (CET) dataset. The contents in red and blue tables would be predicted for conversation completion. The fields in bold are newly introduced by our work in CCET.}
\label{fig:data_structure}
\end{figure*}

\section{CCET Corpus Construction} \label{sect:expt-eval}

In this section, we produce a fuse-motive dataset called Chinese Chat Enhanced Task (CCET) for a generally more functional TOD dialogue system. The corpus is constructed on the basis of three open-source datasets, RisaWOZ, LCCC-base and NaturalConv. We explore the methodology of fusing a new dataset in an automatic process without manual work.

First, to accommodate the TOD type and CC type dialogues to a single paradigm, we redefine the data structure in RisaWOZ dataset to represent the conversational corpus.
Then, an aligning process is employed to annotate the chit-chats dialogue sessions into the new data structure. Finally, a fusing process is performed to re-organize the context parts and the input-output pair parts from the heterogeneous corpora to imitate the user behavior of mode-switching.

Then, a two-stage processing procedure, heterogeneous annotation aligning and context-input fusing, is used for dataset construction. The aligning process is employed to annotate the chit-chats dialogue sessions into the new data structure. 
As a result, the fused CCET dataset could effectively make dialogue models achieve better performances in real-world scenes, where messy CC type user inputs would confuse the systems occasionally.

In this following sections, we explain how to perform the above three procedures in detail.

\subsection{Dialogue Structure Redefinition}
Figure~\ref{fig:data_structure} exhibits a turn-level instance represented using the redefined dialogue structure. It is a TOD type turn from RisaWOZ dataset, in which the active domain switchs from \textit{Attraction} to \textit{Hotel}. In this case, the user has finished asking for information about the attraction \textit{Jinxi Ancient Town} in \textit{Kunshan} region, and tries to find a hotel nearby to live in.

In original annotation of RisaWOZ, the \textit{belief state} (i.e., the slot-value pair \textit{region-Kunshan}) is used to represent user goal at each turn and query a task-specific database (DB). Then, the dialogue policy is determined based on the \textit{Belief State} and the query results. The policy is usually recorded in a structure like ${Domain: Act(slot_1=value_1, slot_2=value_2,...)}$ called \textit{system action} \cite{budzianowski2018multiwoz}, where the Act can be \textit{inform}, \textit{request}, \textit{recommend}, \textit{greeting}, etc. In the instance in Figure ~\ref{fig:data_structure}, the system action is requesting the user to add constraints on hotel rating. Finally, the system action would be converted to a natural language \textit{response} for the user. 

To promote the dialogue model to handle TOD type and CC type interaction in an unified manner, we redefine the data structure in RisaWOZ dataset to depict the conversations in CCET. 
Specifically, one dialogue turn is resolved into three parts, namely \texttt{NLU} (Natural Language Understanding), \texttt{Query}, and \texttt{NLG} (Natural Language Generation).

\paragraph{NLU}
The \texttt{Inputs} to \texttt{NLU} include \textit{Last Belief State}, \textit{Dialogue Context}, and \textit{User Input}. \texttt{NLU} is operated to identify the \textit{Info Msg} and \textit{Query Msg} spans, which respectively model the information provided and requested by user's intents. For instance in Figure~\ref{fig:data_structure}, the two spans are identified as \textit{Hotel:\{region: Kunshan\}} and \textit{Hotel:\{name:?\}}, which represent the intent of finding a hotel in \textit{Kunshan} region.
Specifically, we construct the \textit{Info Msg} span by simply comparing the \textit{Belief State} labels in between turns. The \textit{Query Msg}, which is newly designed in CCET, is constructed grounded in both \textit{belief state} and \textit{system act} labels, involving some work of noisy reducing.
In dialogue models, \textit{Info Msg} updates the \textit{Last Belief State} to construct current \textit{Belief State} to query databases, while \textit{Query Msg} indicates the required information that dialogue systems should provide. 
For chit-chats type users inputs, \textit{Query Msg} is an empty string and systems should generate an interactive open-domain response.

\paragraph{Query}
\texttt{Query} represents the SQL query results, including two parts: \textit{DB Items} and \textit{DB Status}. \textit{DB Items} denotes the exact values about the matching entities, which would appear in the system response. While \textit{DB Status}, by contrast, is an abstraction that represents the bridge between the query results and the basic dialgoue policies. In  Figure~\ref{fig:data_structure} for example, the number of matching hotel entities is more than 10, so the policy should be requesting a slot (e.g. \textit{hotel rating}) for further reducing DB search range.
In CCET, we identify the categories of \textit{DB Status} in accordance with the original RisaWOZ annotations of the number of matched entities and the system actions. In dialogue models, the \textit{DB Status} field would be utilized for policy prediction. 

\paragraph{NLG}
Finally, in \texttt{NLG}, the newly-introduced \textit{Raw Response} denotes the sentence to be generated by conditioning on results of \texttt{NLU} and \texttt{Query}. \textit{Raw Response} has the same meaning of grounded \textit{Final Response}, but contains no variability in terms of linguistic expression, as in Figure~\ref{fig:data_structure}. We use \textit{Raw Response} to replace the \textit{policy} annotation in RisaWOZ so that the demand for supervision data could be declined. We reserve the grounded \textit{Final Response} field as a personalized utterance that might be controllably generated with task-specific designs in future research.

Finally, it is noticed that the original RisaWOZ dataset does not contain \textit{Query Msg}, \textit{DB Status}, and \textit{Raw Response} fields.
As a result, to get these new required information, we pre-define the categories of these data fields with heuristics, and then train several auxiliary classifiers to predict the three fields through an active learning \cite{ren2021survey,simard2017machine} approach as well as a noise reduction work. In CCET, we denote the processed RisaWOZ corpus as \texttt{T}.

\subsection{CC Corpora Alignment}

In this part, we align the dialog sessions from the two chit-chats dialogue corpora, LCCC-base and Neural-Conv, into the pre-defined structure as shown in Figure~\ref{fig:data_structure}. For dialogue systems, it is necessary to distinguish CC inputs from their TOD counterparts. Traditional researches \cite{henderson2014word,zhong2018global,budzianowski2018multiwoz,mrkvsic2016neural} implemented an intent-classification model which infers user's intents based on the current user utterance, where chit-chats was viewed as an additional intent category. Our method, specifically, treats CC type conversation as an out-of-domain request, where no task-oriented information provided or requested by users. In this case, both \textit{Info Msg} and \textit{Query Msg} span are empty strings in the first \texttt{NLU} field, so the input \textit{Last Belief State} always has value \texttt{null} at each turn.
Then, in the second field \texttt{Query}, the \textit{DB Status} is set to a special category \texttt{<NotQuery>}, indicating that system has no need to visit databases at this turn. 
In \texttt{NLG}, the field \textit{Raw Response} is exactly identical to \textit{Final Resp} (i.e. the offered utterance in origin corpora) to sustain the realistic and diverse interaction in open-domain scenario. In CCET, the aligned corpus re-organized from LCCC-base and NaturalConv is denoted as \texttt{C}.

\begin{figure}[bth]
\centerline{\includegraphics[width=\columnwidth] {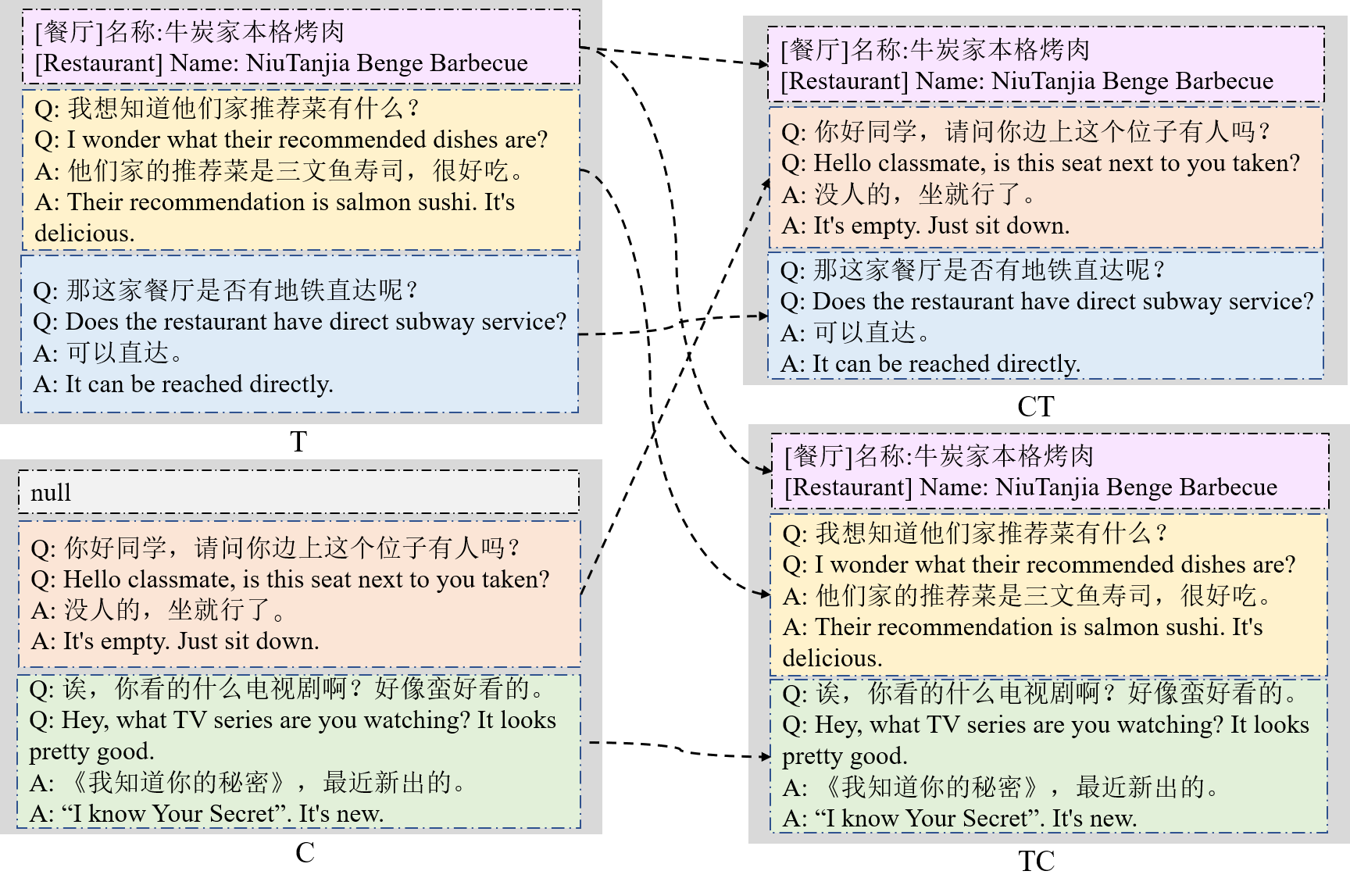}}
\caption{two instances from \texttt{T} and \texttt{C} respectively are joined up to produce new conversational turns. The labels to predict by dialogue systems are omitted for simplicity, and three segments (pink for \textit{last belief state}, orange for \textit{dialogue history}, green for CC and blue for TOD type \textit{current utterances}) are exhibited.}
\label{fig:context_fuse}
\end{figure}

\begin{figure}[htb]

	\quad
	\begin{center}
		\subfigure[A \texttt{T} turn where the user asked for recommendation about a hotel. In this case, new information was informed by user's intent.]{
			\includegraphics[width=6.5cm]{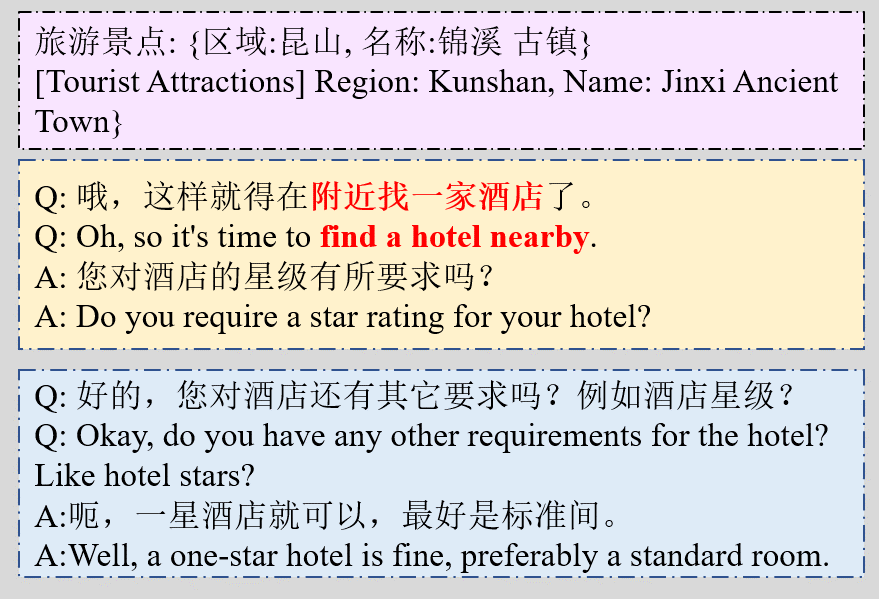}
		}	
	\end{center}
	\quad
	\begin{center}
		\subfigure[A \texttt{T} turn where the system informed a restaurant named \textit{Haoren folk snak}. In this case, the entities changed in the focused restaurant domain.]{
			\includegraphics[width=6.5cm]{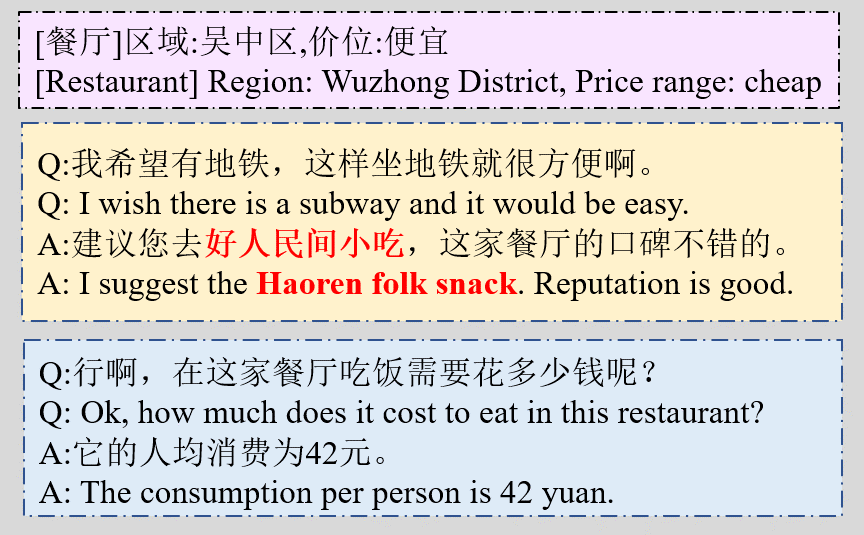}
		}
	\end{center}
\caption{Two \texttt{T} samples where the red spans in bold in dialogue history are semantic conditions for the current utterances essentially. Therefore,  it is  inappropriate to use them for fusion.}
		\label{fig:cccccc}       
\end{figure}

\subsection{Corpora Fusion}

Finally, based on the processed dataset \texttt{T} and \texttt{C}, we fuse the dialogue history part and the input-output pairs part among the two corpora to construct augmented corpora. 
The idea of this step is to mimic the conversational variations in real-life task-completion.

When talking to a real person, the input utterances would be commonly unforeseeable and incoordinate\cite{liu2020robustness,tian2021tod} compared to the elaborately collected corpora, where crowd workers rigidly follow a pre-defined task specification \cite{bohus2009ravenclaw} for interaction. 
In short, the behavioral characteristics of user in practice are inconsistent with the dataset, leading to noisy inputs that models can not handle.

To address this issue, we propose a simple and efficient methodology to imitate the variety in conversational context. In general, a piece of data consists of the dialogue history and current input-output pairs, is denoted as [\texttt{history}, \texttt{current}]. 
In this way, a turn-level sample [$\texttt{history}^\texttt{T}$, $\texttt{current}^\texttt{T}$] in \texttt{T} and another turn-level sample [$\texttt{history}^\texttt{C}$, $\texttt{current}^\texttt{C}$] in \texttt{C} could re-structure with each other to produce new samples [$\texttt{history}^\texttt{T}$, $\texttt{current}^\texttt{C}$] and [$\texttt{history}^\texttt{C}$, $\texttt{current}^\texttt{T}$]. In this way, we fully blend the two kinds of conversations and enrich dialogue diversity.
Figure~\ref{fig:context_fuse} shows one such example of fusing conversational turns. Specifically, the labels to predict by dialogue systems are omitted for simplicity, and three segments (pink for \textit{last belief state}, orange for \textit{dialogue history}, green for CC and blue for TOD type \textit{current utterances}) are exhibited. Among them, contents in orange square are associated with \texttt{history}, while others correspond to \texttt{current}. We fuse the corpora as follows:

\paragraph{Augmentation TC}
the context of \texttt{T} corpus, along with the current input-output pairs of \texttt{C} are joined up to construct a batch of augmented corpus, denoted as \texttt{TC}. It is noticed that we keep the \textit{last belief state} segments in \texttt{TC} unchanged as in \texttt{T}. The reason for this processing is that the \textit{last belief state} segment in \texttt{C} are always \texttt{null} value and irrelevant to the \textit{current utterances} segment. Therefore, we sustain their meaningful counterparts to access linguistic diversity and semantical richness.

\paragraph{Augmentation CT}
The context parts in \texttt{C} corpus, along with the current input-output pairs of \texttt{T}, are joined up to construct a batch of augmented corpus, denoted as \texttt{CT}. Particularly, not all \texttt{CT} results joined in this way are available, as the inter-turn contextual dependency in TOD is stronger in TOD than CC conversation. Therefore, we propose a filtering rule to determine whether an \texttt{T} sample is a suitable material for fusion: 
If (1) the user did not inform new information and (2) all entities in the focused domain kept unchanged in the last turn, 
it means that the original context in \texttt{T} has no information contribution to \texttt{NLU} prediction. In this case, the fusing operation is suitable and the created sample is reserved. Otherwise, the original context in \texttt{T} has strong correlation to current interaction (as the instances shown in Figure~\ref{fig:cccccc}). In this case, the fused result is viewed as invalid and discarded.


\subsection{CCET Corpus Statistics}

The three re-annotated datasets: TOD type RisaWOZ (\texttt{T} in short), CC type LCCCbase (\texttt{C(L)} in short), and CC type NaturalConv (\texttt{C(N)} in short), along with the fused results, form our proposed dataset CCET.
The final constructed dataset contains seven components. The detailed training/testing partition of each component are shown in Table \ref{table:statistics}.
It is noticed that the open-domain CC type corpora, \texttt{C(L)} and \texttt{C(N)}, have much larger size than \texttt{T}. Therefore, taken the size of RisaWOZ as basis, we truncate the conversations from \texttt{C(L)} and \texttt{C(N)} so that the proportions of TOD type and CC type utterances could be relatively balanced. 

While an automatic-produced corpus might have a lower quality than the elaborately-collected human data, it tremendously reduces the potential annotation workload. We advocate that the low quality is acceptable as that the user expressions can be full of chaos when dialogue systems are deployed to users in practice. A training data, which is unclean to some extent, would be able to approximate the realistic data distribution of disfluency.
		
\begin{table}[tbh]
\caption{The number of each segment in CCET dataset.}
\begin{center}
	\begin{tabular}{cccc}
		
		\hline
\multicolumn{2}{c}{Segment}                     & Train & Test \\ \hline
\multicolumn{2}{c}{T}                           & 67290  & 4643   \\ \hline
\multicolumn{1}{c}{\multirow{2}{*}{C}}  & C(N)  & 93390 & 6826    \\ 
\multicolumn{1}{c}{}                    & C(L)  & 93390 & 6827    \\ \hline
\multicolumn{1}{c}{\multirow{2}{*}{TC}} & TC(N) & 67290  & 4643    \\ 
\multicolumn{1}{c}{}                    & TC(L) & 67290  & 46431    \\ \hline
\multicolumn{1}{c}{\multirow{2}{*}{CT}} & CT(N) &55675  & 3709  \\ 
\multicolumn{1}{c}{}                    & CT(L) & 55675  & 3709    \\ \hline
\multicolumn{2}{c}{Total}                       & 500000 & 35000   \\ \hline
		
	\end{tabular}
	
\label{table:statistics}
\end{center}
\end{table}

\section{Fuse-motive Dialogue Framework} \label{sect:expt-evalss}
In this section, we discuss the framework of fused dialogue systems \cite{yan2017building,madotto2020adapter,liu2020robustness} that can handle both TOD and CC dialogue tasks.

\subsection{Task Definition}
Task-oriented dialogue process per turn could be intuitively divided into two sub-tasks, user-side NLU and system-side NLG. 
The NLU module aims to convert the user utterance into the representation that computer can understand,
which includes intent and dialogue act (slot \& value) detection. In this work, NLU is further responsible for extracting dialogue states (user goals) embedded in dialogue context, which is used for searching databases. The NLG module aims to generate a natural language response to the user. 
Some baselines \cite{chen2019semantically,zhang2020task} perform this process by converting the predicted structured system acts into natural language system response. To simplify the system design and reduce the demand of supervision data, we skip the procedure of predicting system acts and directly generate responses by
conditioning on dialogue context and DB status.

\subsection{Evaluation Metrics}

Currently, there remains no standard evaluation methods for fuse-motive dialogue systems so far \cite{deriu2021survey}. In this subsection, we propose several metrics that can be probably available for evaluation on fuse-motive scenarios for dialgoue systems.

\paragraph{TOD metrics}
RisaWOZ itself is a benchmark for TOD evaluation. Although the advantage of fuse-motive models has been mentioned, we still inherit traditional TOD metrics \cite{budzianowski2018multiwoz,lei2018sequicity} to verify that a fuse-motive dialogue system has comparative capability to handle TOD type inputs as traditional TOD baselines \cite{wu2019transferable}. For NLU, the metrics include user action dialog act F1 scores and the Joint Accuracy of DST \cite{zhu2020crosswoz}. As for NLG, only Inform Rate and Success Rate \cite{zhang2020task,wen2016network} are measured and compared with TOD baselines.

\paragraph{CC metrics}
For CC conversation, we follow \cite{adiwardana2020towards} and use PPL (Perplexity) and SSA (Sensibleness and Specificity Average) as the metrics. SSA is a human evaluation metric that computes the average between sensibleness (Does the response
make sense given the context?) and specificity (Is the response specific to the context? ). Moreover, A response can only be considered specific if it is considered sensible.

\paragraph{Mixture metrics}
The key advantage of fuse-motive dialogue models is to response user inputs with both TOD type and CC type capability. The straightforward issue in the mixture of conversation is to decide which capability the user is concerned. Therefore, we simply use dialog act F1 score, a NLU metric in TOD, for evaluation of performances in scenario. Moreover, an additional option is considered: combining all user actions except \texttt{<NotQuery>} together as a single action, then the task would be transformed from a multi classification problem into a binary classification problem.

\subsection{Algorithm Design}
The large size and rich semantic annotations of CCET make it suitable for various kinds of algorithms in NLP. In general, a classic pipelined TOD system including natural language understanding (NLU), dialogue state tracking (DST), dialogue policy learning, and natural language generation (NLG). Among them, several baselines \cite{lee2019convlab,wu2019transferable,mrkvsic2016neural} for NLU and DST could be adapted in CCET benchmark by viewing CCET as a expanded version of TOD corpora. As for NLG, our proposed dataset provides system responses of both CC and TOD type, and thus can be used as a unique benchmark of context-to-text generation task. To this task, several dialogue models \cite{
	lei2018sequicity,zhang2020task} are available to handle the multi-domain response, by encoding dialogue context to decode system response.

\section{Conclusion}
In view of the complex real-life conversational scenario, we propose a simple and cost-effective approach to automatically integrate chit-chats utterances into task-oriented dialogues. Compared with traditional approaches, we creatively propose a new framework of building fuse-motive dialogue systems and related evaluation metrics. The dialogue systems trained in the newly-organized corpora would have capacity to handle mixed user goals including TOD and chit-chats. We hope above works could propel future work on dialogue systems for fuse-motive tasks.

\section*{Acknowledgment}
This work was supported by the National Key R\&D Program of China under Grant 2019YFF0302601.

\bibliography{references}
\bibliographystyle{IEEEtran}

\end{document}